\documentclass[10pt,twocolumn,letterpaper]{article}

\usepackage{cvpr}
\usepackage{times}
\usepackage{epsfig}
\usepackage{graphicx}
\usepackage{amsmath}
\usepackage{amssymb}
\usepackage{booktabs}
\usepackage[numbers,sort,compress]{natbib}

\usepackage[pagebackref=true,breaklinks=true,letterpaper=true,colorlinks,bookmarks=false]{hyperref}

\cvprfinalcopy

\def\cvprPaperID{3785}

\ifcvprfinal\fi

\begin{document}

\title{Low Bandwidth Video-Chat Compression using Deep Generative Models}

\author{Maxime Oquab$^{\star}${\thanks{Contributed equally}}, Pierre Stock$^{\star*}$, Oran Gafni$^{\star}$, Daniel Haziza$^{\star}$, Tao Xu$^\star$, Peizhao Zhang$^\star$, Onur Celebi$^{\star}$,\\ Yana Hasson$^{\dagger}$, Patrick Labatut$^{\star}$, Bobo Bose-Kolanu$^\star$, Thibault Peyronel$^\star$, Camille Couprie$^{\star}$ \\
$^\star$ Facebook, $^\dagger$ INRIA, {\small{work achieved during internship at Facebook AI Research}} \\
{\tt\scriptsize{\{qas, pstock, oran, dhaziza, xutao, stzpz, celebio, plabatut, bobobose, peyronel, coupriec\}@fb.com, yana.hasson@inria.fr}} }
\maketitle
\begin{abstract}
To unlock video chat for hundreds of millions of people hindered by poor connectivity or unaffordable data costs, we propose to authentically reconstruct faces on the receiver's device using facial landmarks extracted at the sender's side and transmitted over the network. In this context, we discuss and evaluate the benefits and disadvantages of several deep adversarial approaches. In particular, we explore quality and bandwidth trade-offs for approaches based on static landmarks, dynamic landmarks or segmentation maps. We design a mobile-compatible architecture based on the first order animation model of Siarohin et al. In addition, we leverage SPADE blocks to refine results in important areas such as the eyes and lips. We compress the networks down to about 3~MB, allowing  models to run in real time on iPhone~8 (CPU). This approach enables video calling at a few kbits per second, an order of magnitude lower than currently available alternatives.
\end{abstract}

\section{Introduction}

For many smartphone users around the world, video-calling remains unavailable or unaffordable. These users are driven out of this fundamental connectivity experience by the prohibitive cost of data plans or because they depend on outdated technologies and infrastructures. For instance, networks might suffer from congestion, poor coverage, power fluctuations and datarate limits -- 2G networks allow for a maximum of 30 kbits/s.
However, with current technologies, an acceptable video-call quality requires at least a stable 200 kbits/s connection.

\begin{figure}
    \centering
    \includegraphics[width=\linewidth]{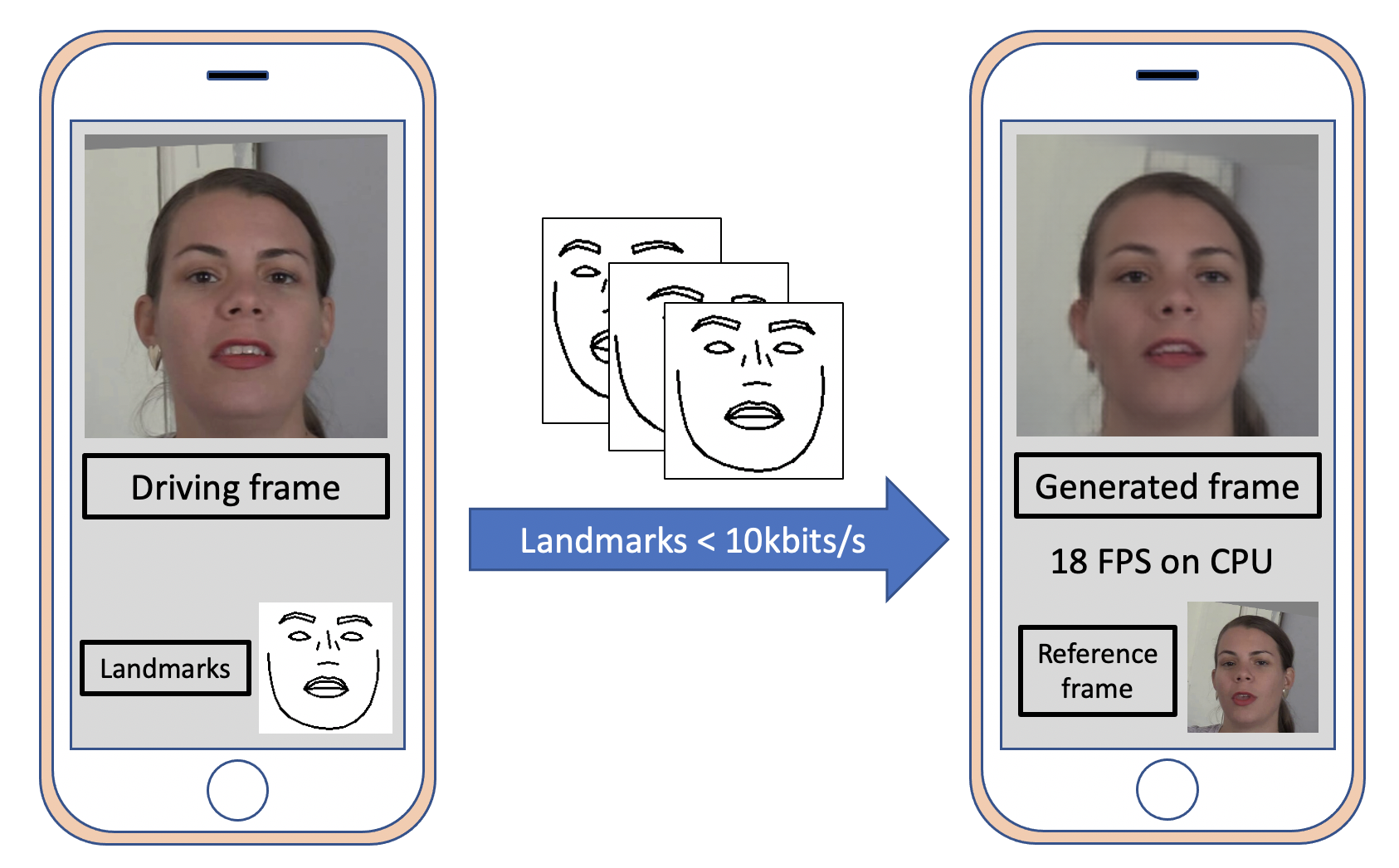}
    \caption{We propose to authentically reconstruct faces in real-time on mobile devices using a stream of compressed facial landmarks extracted from \emph{driving} or {target} frames. The identity of the sender is transmitted in one shot to the receiver at the beginning of the call through a \emph{reference} or \emph{source} frame. This approach is compatible with end-to-end encryption (E2EE).}
    \label{fig:teaser}
\end{figure}

Meanwhile, the research in generative models has now come to a point where the quality of synthetic faces are sometimes indistinguishable from real videos \cite{dolhansky2019deepfake}. To name a few, we may cite Deep video portraits~\cite{Hyeongwoo2018deepvideo}, X2Face~\cite{Wiles_2018_ECCV}, FSGAN~\cite{nirkin2019fsgan}, Neural Talking Heads~\cite{zakharov2019few}, the Bilayer model~\cite{zakharov2020fast} and the First Order Model \cite{Siarohin_2019_NeurIPS}. This unprecedented performance can now be exploited to the benefit of higher quality video calls.
However, there remain important challenges to address before generative models can offer ultra-low data-rate video-calling. In particular, to unlock duplex video-calling for users with last-mile connectivity issues or limited data plans, the models need to be light and fast enough to run on mobile handsets. In addition, to deliver a more seamless and authentic experience, the models should adapt to the current appearance of the user without additional training. In this work, we focus on identifying the best generative strategy compatible with real-time inference on device. We discuss the following approaches:

\begin{itemize}
\itemsep0em
    \item The Neural Talking Heads model~\cite{zakharov2019few}, which requires sending a stream of landmarks in addition to an initial face embedding.
    \item The Bilayer model~\cite{zakharov2020fast}, where the face is reconstructed from a stream of landmarks and a reference frame sent once.
    \item The SegFace model, a novel architecture based on SPADE~\cite{park2019SPADE}, adapted to face animation, which requires sending an initial face embedding and a stream of semantic segmentation maps.
    \item The First Order Model (FOM)~\cite{Siarohin_2019_NeurIPS}, which requires sending ten landmarks, their associated motion matrices, and one reference frame.
\end{itemize}
Analysing the FOM in depth, we observe that only sending the landmarks compressed with Huffman coding (no motion matrices) achieves sufficient quality and leads to an outstanding data-rate reduction. Compared to other approaches, this model allows for good identity and background preservation. Our contributions are the following:
\begin{itemize}
\itemsep0em
    \item We provide a comparative analysis of leading generative approaches for the specific use-case of enabling ultra-low data-rate video calling.
    \item We develop a strong baseline leveraging the SPADE architecture and segmentation maps.
    \item We propose a warping based approach leveraging SPADE blocks to refine important face attributes such as eyes and lips.
    \item While previous approaches were tested on specialized hardware (servers, mobile GPU), we provide first real-time results on mobile CPU.
\end{itemize}

\section{Related work}

\subsection{Face compression before deep learning}

The idea of face-specific video compression is not novel and appeared with classical computer vision tools, for instance morphings using Delaunay triangulations, Eigenfaces, or 3D models.
The first reference we found on the topic is the work of Lopez et al.~\cite{lopez1995head} that proposes  to encode only pose parameters of a 3D head model, which is projected to reproduce a video sequence.

Previous work~\cite{koufakis1999very} use PCA to model the current frame as a linear combination of three basis frames sent prior to the call. The authors rely on known control points  on the face boundaries and landmarks. The principal drawback of the approach is the presence of triangulation artefacts, even when a large number of control points is used. The achieved bandwidth is 1500 bits/frame.
Similar usage of Eigenspaces are suggested in \cite{tuceryan2000model,torres2002proposal,soderstrom2006very}.
Among these proposals using Eigenspaces, one claims an extremely low bit-rates achievement of 100 bits/s \cite{son2006ultra}. However the proposed solution is hard to scale, as it requires storing personal galleries of face images to reconstruct videos at the receiver side.

\subsection{Deep compression}

The emergence of Generative Adversarial Networks (GANs) stimulated the application of deep learning to video compression. Super-resolution has been an active field of research leveraging GANs for image and video compression. There have been a number of research works tackling this problem \cite{chen2018fsrnet,Ustinova2017facehallucination,bulat2018super}. However, for compressing faces, these reconstructions methods are limited to restoring personal traits from low level images and only work well for limited upscaling factors (around $2\times$ in resolution).
The power of GANs for lossy image compression started to be demonstrated in the
Generative compression work of Santukar et al.~\cite{Santurkar2017generative}, using an auto-encoder combined with adversarial training. The state-of-the-art has since improved with the Extreme Learned Image Compression work of Agustsson et al.~\cite{agustsson2019generative}, thanks to a multi-scale architecture and the usage of semantic segmentation information, among other tricks used by the authors.
The work of Liu et al.~\cite{liu2020deep} surveys deep learning-based approaches for general purpose video compression. Among them, Learned Video Compression \cite{waveonevideo2018} demonstrates for the first time the superior capacity of an end-to-end machine learning approach over standard codecs.
By focusing on faces only, we can lower the bandwidth, improve the quality and compress models compared to using more generic methods. Therefore, we review next deep face videos reconstruction approaches and their adequacy to video chat compression.

\subsection{Deep talking head approaches}

3D based approaches produce realistic avatars which can be animated in real-time~\cite{Cao2016RealtimeFacial}. However, such methods require to capture a set of images of the user (a few dozens) to build their personal face model. PAGAN \cite{nagano2018pagan} generates key face expression textures that can be deformed and blended in real-time on mobile from a single frame. However, the reconstruction of certain features, notably the hair, is still problematic in 3D model-based approaches. Deep video portraits \cite{Hyeongwoo2018deepvideo} is handling this issue using a rendering-to-video translation network, but the approach needs about a thousand images per subject for training.
Stimulated by advancements in face swapping pipelines \cite{korshunova2017fast, Wiles_2018_ECCV}, a number of deep generative re-enactment approaches arose. Contrary to warping based re-enactment~\cite{averbuch2017bringing}, learning faces reconstructions enables extra robustness in presence of large head angles.
The Face Swapping GAN \cite{nirkin2019fsgan} relies on several steps: landmarks extraction, segmentation, interpolation and inpainting. This complex pipeline may result in robustness issues and limited bandwidth gain due to the need of sending both compressed segmentations and landmarks. Similarly, the vid2vid approach (\cite{wang2018vid2vid}, \cite{wang2019fewshotvid2vid}) requires sending a ``sketch" (edge map) for each frame in order to re-enact a face, which has a relatively high bandwidth cost.
In the next section, we discuss and compare the learning based approaches which yield the most promising results in terms of bandwidth, visual quality and inference time.

\begin{figure*}
\includegraphics[width=\linewidth, trim=0cm 5.7cm 1.5cm 2.1cm, clip=true]{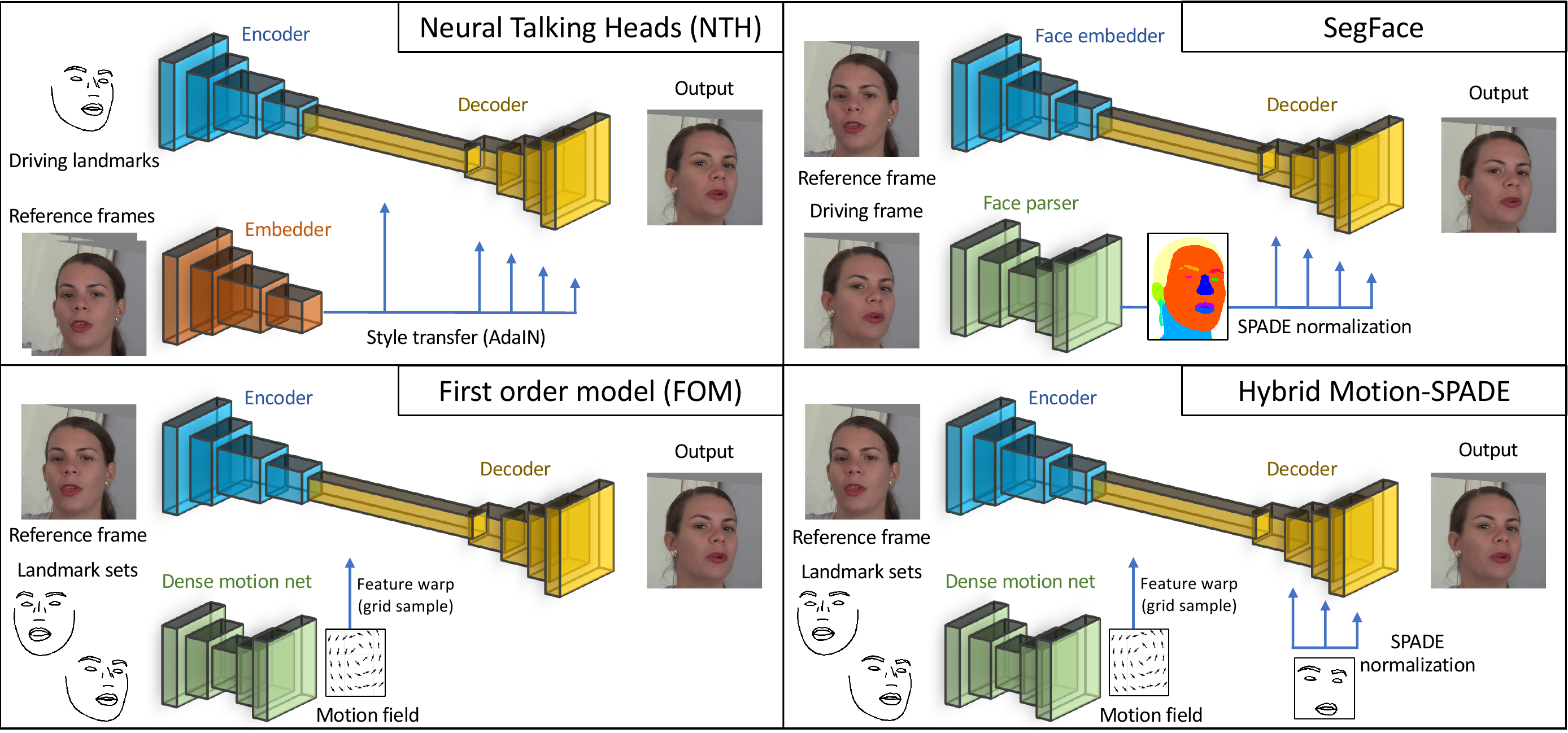}
    \caption{Scheme of principle for the different deep generative approaches discussed in this study. In particular, we detail two novel architectures, SegFace and Hybrid Motion-SPADE (right) and compare them to existing NTH~\cite{zakharov2019few} and FOM~\cite{Siarohin_2019_NeurIPS} models (left). For all models, we assume the generation is performed by the encoder-decoder pair on the receiver device, while the emitter sends a reference frame (or several) at the beginning of inference, and streams a series of landmarks or segmentation maps.}
    \label{fig:models}
\end{figure*}

\section{Generative models}

In this section we describe in-depth several recent face animation algorithms that we have implemented and studied.  We share our understanding of these works and present our two model contributions, namely SegFace and Hybrid Motion-SPADE. An overview of these different models appears in Fig.~\ref{fig:models}.
For this self-reenactment task, unless mentioned otherwise, the goal of all these approaches is to generate a frame based on (i) one fixed \emph{reference} or \emph{source} frame and (ii) position information (e.g. landmarks) from a stream of  \emph{driving} or \emph{target} frames (see Figure~\ref{fig:teaser}).

\subsection{Talking heads (NTH) and Bilayer model}

The “Talking Heads” work of Zakharov et al.~\cite{zakharov2019few} learns to synthesize videos of people from facial landmarks given one or few reference images. It follows an encoder-decoder architecture with a style transfer component:
\begin{itemize}
\itemsep0em
    \item A set of style parameters is computed for the set of reference images.
    \item Facial landmarks are plotted as images and processed by an encoder network.
    \item The resulting code is decoded with style transfer, using Adaptive Instance Normalization \cite{Huang_2017_ICCV} layers, adjusting the mean and standard deviation of each feature map with the style parameters.
\end{itemize}

The networks are trained end-to-end with adversarial and perceptual losses on a dataset of videos. The best results are achieved by performing a fine-tuning training phase on the generator to match the reference frames as precisely as possible. This fine-tuning phase requires several minutes on a modern server GPU. Without fine-tuning, the identity is not preserved as well in the generated frames. In practice, a few hundreds of frames would have to be sent at the beginning of the call.

This work was further improved in the Bilayer Synthesis approach~\cite{zakharov2020fast}, where the fine-tuning step is not required anymore, and leads to visually appealing and sharp results. In our observations (see Figure \ref{fig:comp}), the identity preservation suffers from a stronger uncanny valley effect.

In terms of bandwidth, the NTH and Bilayer approaches require sending 68 compressed landmarks.

\subsection{First order model for image animation (FOM)}

The ``First Order Model" approach of Siarohin et al.~\cite{Siarohin_2019_NeurIPS} deforms a reference source frame to follow the motion of a driving video. While this method works on various types of videos (Tai-chi, cartoons), we focus here on the face animation application. FOM follows an encoder-decoder architecture with a motion transfer component:

\begin{itemize}
\itemsep0em
    \item A landmark extractor is learned using an equivariant loss, without explicit labels.
    \item Two sets of ten learned landmarks are computed for the source and driving frames.
    \item A dense motion network uses the landmarks and the source frame to produce a dense motion field and an occlusion map.
    \item The encoder encodes the source frame.
    \item The resulting feature map is warped using the dense motion field (using a differentiable grid-sample operation \cite{jaderberg2015spatial}), then multiplied with the occlusion map.
    \item The decoder generates an image from the warped map.
\end{itemize}

The networks are trained end-to-end on video frames, using perceptual losses, and are then optionally fine-tuned with an adversarial discriminator. The self-supervised landmarks do not necessarily match precise locations of the face. Instead, they correspond to point coordinates that are optimized to achieve the best deformation of the source frame. \cite{Siarohin_2019_NeurIPS} describes how to improve motion approximation in landmark areas by estimating Jacobian matrices to model motion in their neighborhood.
In our observations (see Table \ref{tab:humanStudy}), this approach preserves identities better than NTH and is at least on par with the follow-up Bilayer synthesis approach.
Next, we study variants of this approach.

\paragraph{Variants}
First, our implementation does not use the Jacobian component, as we do not observe a strong effect the quality of the results. We refer to the resulting model as ``Motion Net (MN-10)" as it no longer uses first order approximation anymore and employs a set of ten landmarks.

Second, we explore using off-the-shelf facial landmarks extraction to replace the unsupervised landmarks. In this case, we only stream 20 or 68 compressed landmarks.

Third, we explore a combined strategy employing both 10 self-supervised landmarks and 20 supervised ones, that we note MN-10+20. We introduce a fourth variant in Section~\ref{subsec:hybrid_motion_spade}, after detailing our SegFace approach below.

\subsection{SegFace}

This approach builds upon \cite{park2019SPADE}. Unlike MaskGAN~\cite{lee2020maskgan}, we propose to use a face descriptor computed on a source frame, and decode it by conditioning on face segmentation maps from a driving frame. It follows an encoder-decoder architecture described as follows:

\begin{itemize}
\itemsep0em
    \item A face descriptor is computed on a source frame.
    \item This face descriptor is given to a decoder network, that applies SPADE normalization blocks at each layer using the face segmentation maps of the driving frame, ensuring all parts of the face are correctly placed.
\end{itemize}
The decoder network is trained using VGGFace2 face embeddings \cite{cao2018vggface2}, and segmentation maps from \cite{yu2018bisenet}
as inputs. Its objective during training is to reconstruct the same source frame. The optimization is done using losses from \cite{park2019SPADE}, and the face perceptual loss from \cite{gafni2019live}. This method operates on independent frames, and thus allows to use high-resolution training data, leading to high quality results. Training is achieved using CelebA \cite{liu2015faceattributes} and Flickr-Faces-HQ datasets.

\paragraph{Bandwidth} The model requires a segmentation map labeled for 15 categories (eyes, hairs, ears etc.). Sending compressed segmentation maps would require 18/25 kbits/s at resolutions $48\times$/$64\times$, knowing that there is a trade-off between the resolution of the transmitted segmentation maps and the quality of the generated faces.
We do not build on this method further for low-bandwidth video-chat because the cost of running a face parser inference step and the bandwidth requirements are too high. The SegFace implementation, however, allows us to observe that the generated images respect the segmentation map labels almost perfectly, consistently with the conclusions of \cite{lee2020maskgan}. We will build on this property in the next subsection with our Hybrid Motion-SPADE approach.

\subsection{Hybrid Motion-SPADE model}
\label{subsec:hybrid_motion_spade}

 Important quality criteria for compressed video-chat include a good synchronization between the lips and the speech, and a good rendering of the eyes and eyebrows; therefore, it is crucial to generate these facial parts precisely.

 We propose an improvement over the FOM-based Motion Net method, by adding SPADE normalization layers in the upsampling blocks of the decoder network (in the last step of the FOM approach). We draw polygons for the eyes, eyebrows, lips and inner mouth using 60 extracted face landmarks, and use these as semantic maps for SPADE.

The dense motion network receives (i) a downsampled reference frame with (ii) the positions of $N$ landmarks for that frame, and (iii) the positions of the same landmarks for a driving frame.
It outputs a motion field $M$ and an occlusion map $O$.
The encoder network outputs a feature map $F_s$.
The decoder warps $F_s$ with the result of the dense motion network $M$ and multiplies it element-wise with the occlusion map $O$, to obtain $F_w$.
Then, $F_w$ is processed by a stack of five residual blocks and three upsampling blocks that apply the SPADE normalization using a set of 60 landmarks.

Training is performed with a multiscale perceptual loss (based on a VGG-19 architecture) with a weight $\lambda_p=10$ in addition to an equivariance loss with a weight $\lambda_{eq}=1$ for the unsupervised landmark detector when applicable, following the procedure described in \cite{Siarohin_2019_NeurIPS}.

\paragraph{Bandwidth}
The necessary segmentation maps are obtained by plotting the polygons of the facial landmarks extracted using a landmark detector (see Figure \ref{fig:models}), rather than running a face segmentation network. Moreover, landmark coordinates are inexpensive to transmit, while rasterized segmentation maps are more difficult to compress, especially at higher resolutions.
In terms of bandwidth, this approach requires sending $N+60$ compressed landmarks. We experiment with $N=10, 20,$ and 30.

\section{Compression}

\begin{table*}[]
    \centering
    \begin{tabular}{ll|c|ccc|cc}
    \toprule
         Model variant & Inputs & FPS & \#Params & \#FLOPS & int8 size & Raw BW & Compressed BW \\
         \midrule
         Motion Net & 10 U & 18 & 2.9 M & 1411 M & 3.1 MB & 3.9 kbits/s & 1.4 kbits/s \\
         Motion Net & 20 L & 19 & 2.3 M & 1293 M & 2.5 MB & 7.8 kbits/s & 2.2 kbits/s \\
         Motion Net & 10 U + 20 L & 14 & 3.0 M & 1505 M & 3.4 MB & 11.7 kbits/s & 3.6 kbits/s \\
         Motion SPADE & 10 U & 16 & 2.9 M & 1198 M & 3.2 MB & 27.3 kbits/s & 8.0 kbits/s \\
         Motion SPADE & 20 L & 19 & 2.3 M & 1029 M & 2.5 MB & 41.2 kbits/s & 8.8 kbits/s \\
         Motion SPADE & 10 U + 20 L & 13 & 3.0 M & 1292 M & 3.4 MB & 35.3 kbits/s & 10.2 kbits/s \\
         \bottomrule
    \end{tabular}
    \vspace{1ex}
    \caption{Comparison of our approaches running on mobile in terms of compression for both model size and stream.``10 U" (resp. ``20 L") means that 10 unsupervised landmarks (resp. 20 facial landmarks) are used as inputs to the dense motion network. SPADE variants require 60 extra facial landmarks to draw the facial label maps. Notes: the ``int8 size" is the full combined size of the models. The number of frames per second (FPS) is measured for the whole int8-quantized pipeline running on an iPhone~8, including landmark detection,  grid-samples and face alignment. The \#FLOPS count is for the dense motion, decoder, and unsupervised landmark extractor networks. The bandwidth (BW) is measured at 25 FPS, without (Raw BW) and with Huffman encoding (Compressed BW).}
    \label{tab:size_compressed}
\end{table*}

In this section, we explain different strategies to make architectures -- and in particular our novel hybrid Motion-SPADE -- compatible with low-bandwidth video calls on mobile. We first detail the architectures and then the compression aspects for the models and the bandwidth. Results are displayed in Table~\ref{tab:size_compressed}.

\subsection{Mobile architectures}

\paragraph{Base blocks}
We rely on the open-source FbNet family of architectures~\cite{wu2019fbnet,wan2020fbnetv2,dai2020fbnetv3} to design mobile-capable models for our Motion Net and Motion-SPADE approaches. These networks typically build on blocks combining $1\times1$ point-wise and $3\times3$ depth-wise convolutions~\cite{sandler2018mobilenetv2}  that require less floating-point operations than traditional $3\times 3$ convolutions found in residual blocks. We provide further architecture details in Figure \ref{fig:FOM_scheme2}.

\paragraph{Mobile SPADE normalization blocks}
When applicable, we perform a SPADE normalization after the last $1\times1$ point-wise convolution, with kernel sizes of $1\times1$, and 32 hidden channels. We have found these parameters to provide a good trade-off between speed and quality while preserving the fidelity of the SPADE approach.

\subsection{Landmark stream compression}

We compress the landmarks with Huffman encoding~\cite{huff}. In this approach, the landmark displacements are first binarized into 32 bins plus one sign bit, and we encode the bin index with Huffman coding.  This compression leads to an average rate of 90 bits/frame for 20 landmarks, hence 2.2 kbits/s at 25 FPS (see Table~\ref{tab:size_compressed} for details). For reference, bandwidth requirements for audio are around 10 kbits/s, while the AV1 video codec (not widely hardware-supported to date) aims at 30~kbits/s~\cite{bloggoogle}. Therefore, we did not explore other variants such as Arithmetic Coding \cite{arith} since the audio part takes most of the bandwidth of a call with the proposed approach.

\subsection{Model quantization}

We rely on \texttt{int8} post-training quantization. This technique simply consists in uniformly quantizing both weights and activations over 8 bits, thus reducing the model size by a factor $4$. Moreover, \texttt{int8} models traditionally benefit from a $\times 2-3$ speed-up compared to their \texttt{fp32} counterparts for both server and mobile CPUs. The scale and zero-point parameters\footnote{The affine transform coefficients that allow converting an 8-bit quantized tensor (integer-valued in $[0,255]$) to its floating-point counterpart.} of the quantized layers are calibrated after training using a few batches of training data. When not properly calibrated, we found that the decoder generates an image with a small amount of grain or noise, resulting in a loss of visual quality.

To compress the Motion based models, we only rely on \texttt{int8} since the non-compressed models are already small. The models are converted to TorchScript and run on the phone's CPU. These results are displayed in Table~\ref{tab:size_compressed}.

\subsection{Implementation details}

Our mobile models are trained on the DFDC dataset \cite{dolhansky2019deepfake} rather than the VoxCeleb \cite{Nagrani17vox} dataset, in contrast to the original work of \cite{Siarohin_2019_NeurIPS} (though we provide evaluation numbers for comparison and reference).
We split different identities following a 90\%-10\% ratio, resulting in a total of 21899 training videos, and 2369 validation videos.
We choose DFDC in this work because the videos are higher-quality and not cropped as tight, allowing for different face alignment procedures: (i) cropping around the face, or (ii) cropping after rotation using the facial landmarks such that the eyes are horizontally aligned. We have notably found that for smaller Motion Net models, this alignment makes the task easier and improves the results. The alignment procedure is reproduced on mobile at inference time to match the training distribution.

We perform training on 8 GPUs using a Distributed Data Parallel pipeline in Pytorch, with a batch size of 48, for 265K steps. We use the Adam optimizer with learning rate $2.10^{-4}$ on all networks in all experiments.

\section{Experiments}

\subsection{Evaluation metrics}

\begin{figure*}[htb]
    \begin{tabular}{ccccccc}
     \includegraphics[width=0.12\linewidth]{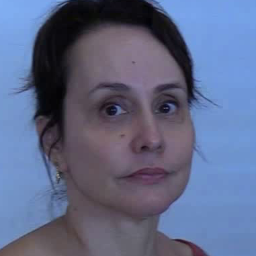}&
    \includegraphics[width=0.12\linewidth]{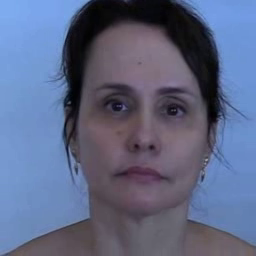}&
    \includegraphics[width=0.12\linewidth]{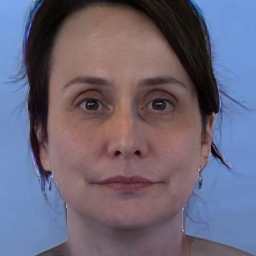}&
    \includegraphics[width=0.12\linewidth]{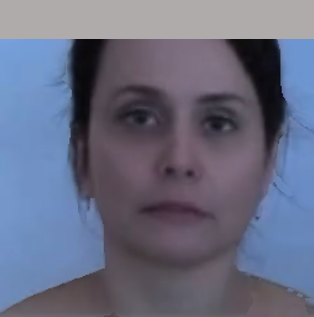}&
    \includegraphics[width=0.12\linewidth]{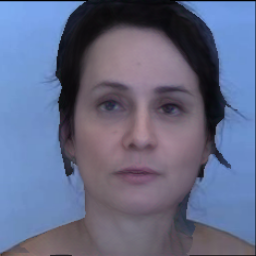}&
     \includegraphics[width=0.12\linewidth]{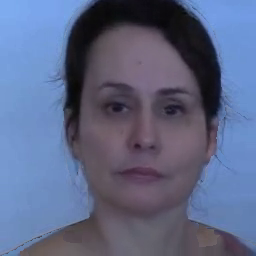} &
    \includegraphics[width=0.12\linewidth]{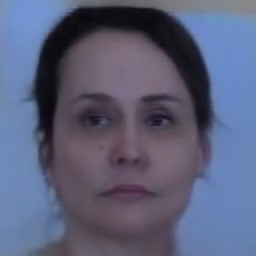}\\
     \includegraphics[width=0.12\linewidth]{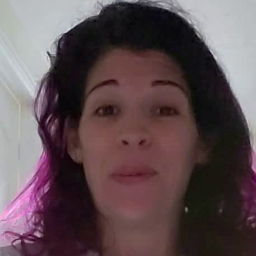}&
    \includegraphics[width=0.12\linewidth]{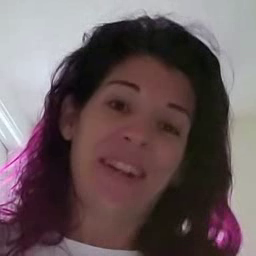}&
    \includegraphics[width=0.12\linewidth]{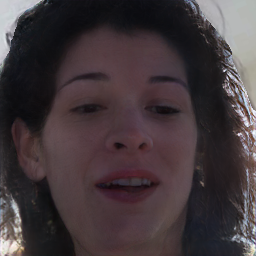}&
    \includegraphics[width=0.12\linewidth]{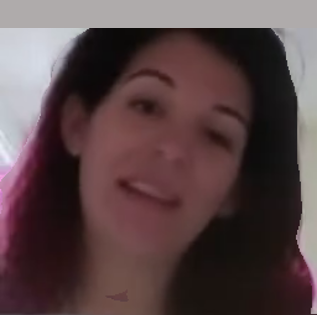}&
    \includegraphics[width=0.12\linewidth]{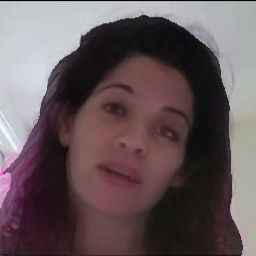}&
     \includegraphics[width=0.12\linewidth]{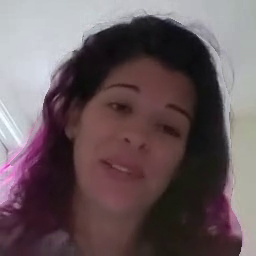} &
    \includegraphics[width=0.12\linewidth]{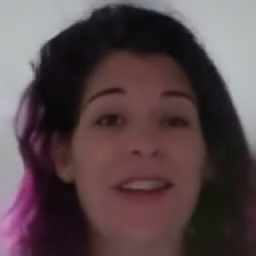}\\
     \includegraphics[width=0.12\linewidth]{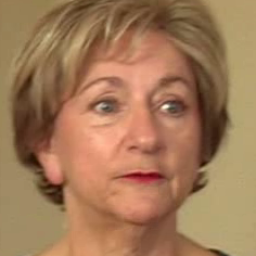}&
    \includegraphics[width=0.12\linewidth]{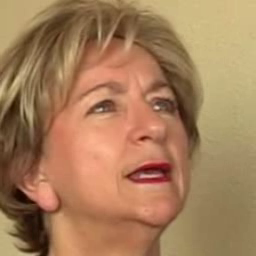}&
    \includegraphics[width=0.12\linewidth]{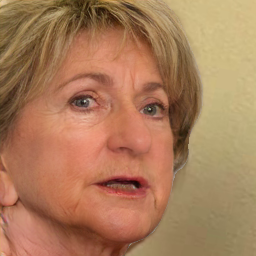}&
    \includegraphics[width=0.12\linewidth]{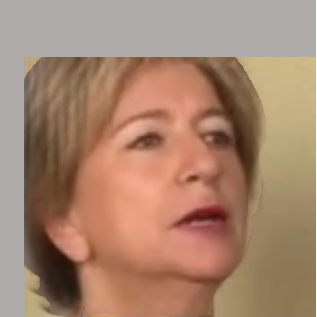}&
    \includegraphics[width=0.12\linewidth]{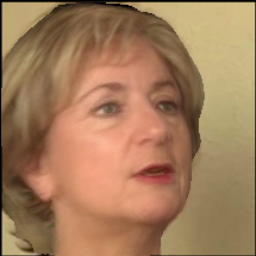}&
     \includegraphics[width=0.12\linewidth]{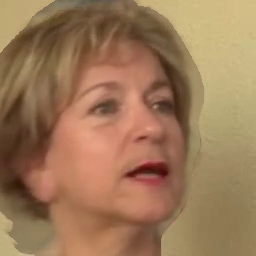} &
    \includegraphics[width=0.12\linewidth]{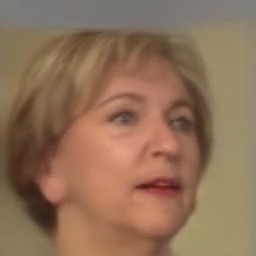}\\
     \includegraphics[width=0.12\linewidth]{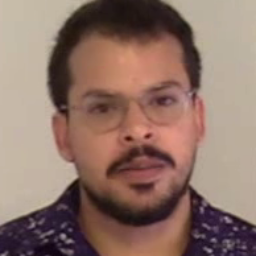}&
    \includegraphics[width=0.12\linewidth]{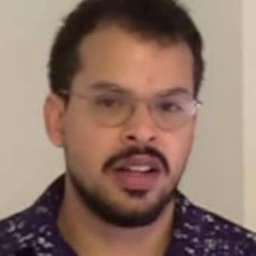}&
    \includegraphics[width=0.12\linewidth]{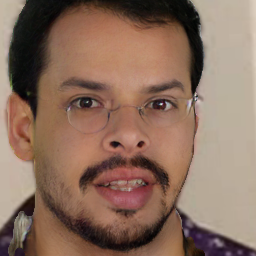}&
    \includegraphics[width=0.12\linewidth]{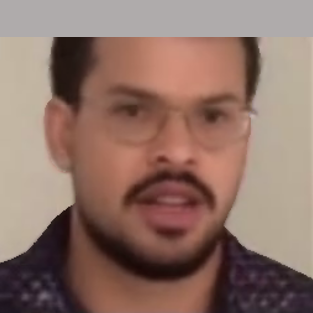}&
    \includegraphics[width=0.12\linewidth]{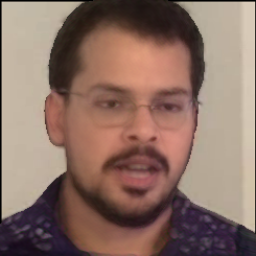}&
     \includegraphics[width=0.12\linewidth]{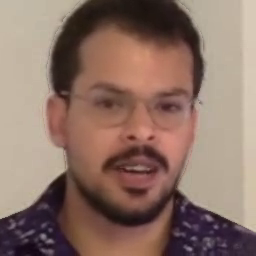} &
    \includegraphics[width=0.12\linewidth]{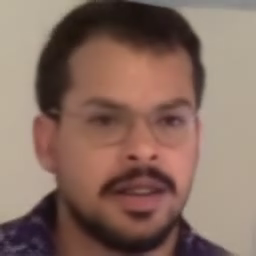}\\
     Source &  Target & Seg2Face &  NTH          & Bilayer   &  FOM adv & Mob MS-20L\\
                   &   & \#param: 126M & \#param: 34M & \#param: 144M & \#param: 47M & \#param:3M \\
    \end{tabular}
    \vspace{2ex}
    \caption{Comparison of different results using  Seg2Face ($48\times48$), NTH, Bilayer, and FOM adv. Each model generates the face using the fixed source frame and the facial information (such as landmarks) of the driving frame. We pasted ground truth backgrounds to have a fair evaluation. The last column showcases the results using our Mobile Motion-SPADE that runs at 18 FPS on an iPhone 8, whereas the other models run on server, have at least 10$\times$ more parameters and are not necessarily compatible with low-bandwidth video calling. Note that the alignment procedure may differ between the models, hence the head is not centered the same on the generated faces.}
    \label{fig:comp}
\end{figure*}

We evaluate the models using the perceptual LPIPS \cite{zhang2018unreasonable} and multi-scale LPIPS-like metrics employed in \cite{Siarohin_2019_NeurIPS}, that we name msVGG.
Second, as argued in \cite{chen2020comprises}, the cosine similarity CSIM computed between features of the pre-trained face embedding network ArcFace \cite{deng2019arcface} is one of the most effective metric to assess quality of talking heads models, we therefore report it.
Finally, we quantify facial landmarks mismatch by running a landmark detector on the true and generated videos and computing the Mean Square Error between each pair of landmarks. This metric is classically referred to as the Normalized Mean Error (NME) of head pose \cite{bulat2017far}.
All the generative approaches considered in this work are trained using different alignments and close-ups (see Figure~\ref{fig:comp}), so we compute our metrics using the optimal modified videos for each method as ground truth.

\subsection{Quality evaluation: ablation studies}

 For evaluation, we assembled a set of 28 videos of diverse persons in terms of gender, age, skin color from the validation set of VoxCeleb2 \cite{Chung18bvox2}, and a similar set of 50 videos from the validation split of the DFDC dataset~\cite{dolhansky2019deepfake}.

We begin our analysis of the FOM by computing the quality of reconstruction without first order motion approximation and without adversarial training in Table \ref{tab:QualityAblationFOM}. While it is clear that the adversarial fine-tuning boosts the performance, we experiment without it in the remaining of our ablation study around this model to reduce training time for each model. Removing the first order approximation only slightly degrades the LPIPS but not the msVGG perceptual metric. Interestingly, the CSIM metric which is the one supposed to best reflect the identify preservation, is slightly increased by dropping this component. A second observation is that the fidelity of facial landmarks to the target video is negatively affected by this removal.
Since the drop of performance induced by discarding first order motion approximation leads to important bandwidth savings and limited loss in performance, we conduct our experiments without it. We refer to this approach as the Motion Net approach.
Next, we explore the replacement of the self-supervised landmarks of the Motion Net approach by off-the-shelf landmarks from a state-of-the art detector. Results appear in Table \ref{tab:QualityAblationMN}.
Note that the results presented in this table are obtained by our re-implementation of the MotionNet approach, and are slightly better than these of Table~\ref{tab:QualityAblationFOM} obtained with the original code.
\begin{table}[]
    \centering
    \begin{tabular}{cccc}
    \toprule
         & FOM adv & FOM w/o adv  & MN\\
         \midrule
        msVGG $\downarrow$ & {\bf 85.6} & 87.5 & 87.9 \\
        LPIPS $\downarrow$& {\bf 0.226} & 0.233 & 0.236\\
        NME $\downarrow$& {\bf 0.51} & 0.53 & 0.54\\
        CSIM $\uparrow$& {\bf 0.83} & 0.81 & 0.82\\
        \bottomrule\\
    \end{tabular}
    \caption{Ablation study for FOM on VoxCeleb2-28. MN: FOM without first order approximation nor adversarial fine-tuning.}
    \label{tab:QualityAblationFOM}
    \vspace{-2ex}
\end{table}
We compare in Table~\ref{tab:QualityAblationMN} different variants of the Motion Net approach, using
20 input landmarks, 68 input landmarks, self-supervised landmarks with dense architectures and with mobile architectures. All these dense architectures employ a latent space of $256\times64\times64$, and were trained on VoxCeleb.
Using standard facial landmarks instead of unsupervised motion landmarks degrades the scores of perceptual metrics, but improves NME. Using 68 landmarks is only very slightly improving the quality over 20. With mobile architectures, we reduce the latent space to $256\times32\times32$. In addition to using 10 motion landmarks or 20 landmarks alone, combining these two sets helps boost all quality metrics. Finally, we observe that adding the SPADE blocks preserves the perceptual quality and brings a large improvement in NME.

\begin{table}[]
    \centering
    \begin{tabular}{lccc}
    \toprule
      & \small LPIPS $\downarrow$ & \small NME $\downarrow$ & \small CSIM $\uparrow$ \\
    \midrule
          \small Dense MN-10 U & 0.221 & 0.59 & 0.83\\
          \small Dense MN-20 L & 0.242 & 0.50 & 0.80\\
          \small Dense MN-68 L & 0.240 & 0.49 & 0.81\\
          \midrule
          \small Mob MN-10 U & 0.225 & 0.52 & 0.79 \\
          \small Mob MN-20 L & 0.244 & 0.48 & 0.78 \\
          \small Mob MN-10 U + 20 L &  0.218 & 0.46 & 0.80 \\
          \small Mob M-SPADE-10 U &  0.217 & 0.47 & {\bf 0.81} \\
          \small Mob M-SPADE-20 L & 0.242 & {\bf 0.44} & 0.79 \\
          \small Mob M-SPADE-10 U + 20 L & {\bf 0.215} & 0.46 & {\bf 0.81} \\
        \bottomrule \\
    \end{tabular}
    \caption{Evaluation results for Motion Net approaches without adversarial fine-tuning on the VoxCeleb2-28 video subset. Mob : Mobile models. Dense models ($64\times64$ latent space) are trained on VoxCeleb. Mobile models ($32\times32$) are trained on the DFDC aligned dataset. U: unsupervised landmarks; L: facial landmarks.}
    \label{tab:QualityAblationMN}
\end{table}

\subsection{Quantitative comparative evaluation}

We compare the quality/bandwidth trade-off of different dense face animation approaches in Table~\ref{tab:QualityBandwidth}.
As the different models were trained using different data pre-processing (different crops, alignment), we evaluate each one in the setting allowing the best generations. This means that synthesized videos need to be compared to different source videos.
 Therefore, we paste the ground truth video background on the generated result so that the metrics focus on evaluating face differences only.
We observe that the NTH results lead to better NME, and FOM to better LPIPS and CSIM metrics. SegFace numerical results lower, partly due to a color shift appearing in the results. Interestingly, our mobile results have better NME and msVGG scores than the original FOM dense approach.

\begin{table}[]
    \centering
    \begin{tabular}{cccccc}
    \toprule
         &\small NTH&\small Bilayer&\small SegFace&\small FOM&\small MS20L\\
         \midrule
        \small msVGG* $\downarrow$ & {\bf 56.3} &  68.6 & 84.4 &  58.7 & 57.9\\
        \small LPIPS* $\downarrow$ & 0.165      & 0.200 & 0.304 & {\bf 0.153} & {0.167}\\
        \small NME* $\downarrow$ & {\bf 0.38}   & 0.55  & 0.55 &  0.50 & 0.44\\
        \small CSIM* $\uparrow$ & 0.83          & 0.85  & 0.76 & {\bf 0.87} & 0.84\\
        \small kbits/s $\downarrow$  & $9.7^+$      & 9.7   & 18 & {\bf 4.0} & 8.8 \\

        \bottomrule\\
    \end{tabular}
    \caption{Comparison of Bilayer, SegFace ($48\times48$),
     FOM adv, NTH in terms of quality / bandwidth (kbits/s with 25 fps) trade-offs on VoxCeleb2-28. * : Metrics were computed using ground truth backgrounds. We also include our best mobile model, Motion-SPADE-20L (MS20L) in the comparison. $^+$: Bandwidth computed without considering the mandatory frames transmission necessary to the fine-tuning step.}
    \label{tab:QualityBandwidth}
\end{table}

\begin{figure*}[htb]
      \includegraphics[width=0.105\linewidth]{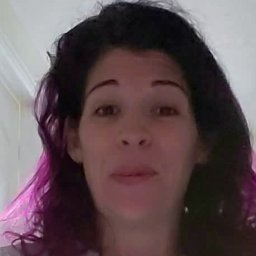}
    \includegraphics[width=0.105\linewidth]{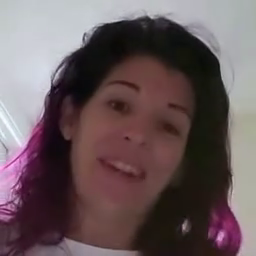}
    \includegraphics[width=0.105\linewidth]{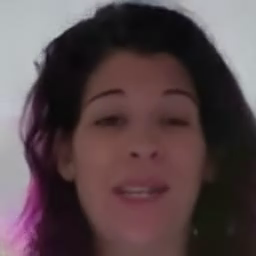}
     \includegraphics[width=0.105\linewidth]{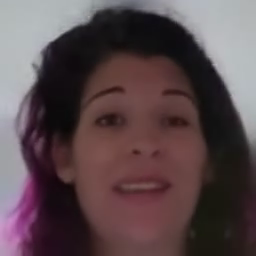}
      \includegraphics[width=0.105\linewidth]{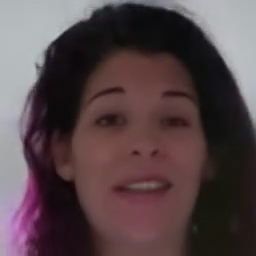}
      \includegraphics[width=0.105\linewidth]{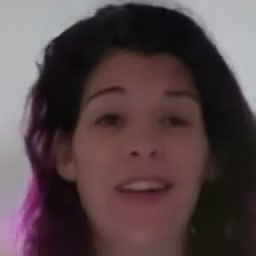}
       \includegraphics[width=0.105\linewidth]{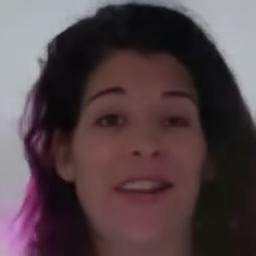}
         \includegraphics[width=0.105\linewidth]{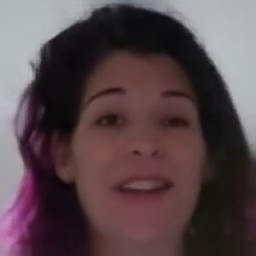}
     \includegraphics[width=0.105\linewidth]{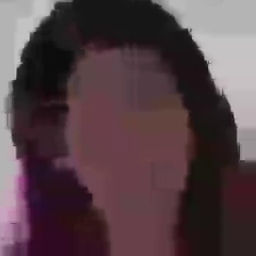}\\
      \includegraphics[width=0.105\linewidth]{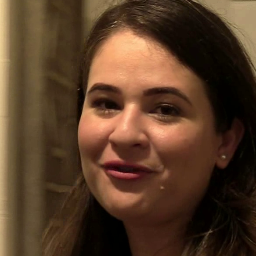}
    \includegraphics[width=0.105\linewidth]{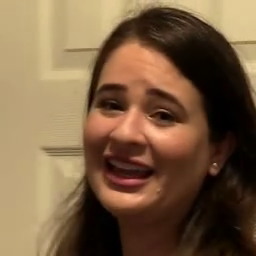}
    \includegraphics[width=0.105\linewidth]{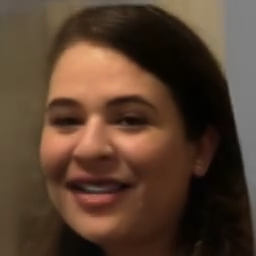}
     \includegraphics[width=0.105\linewidth]{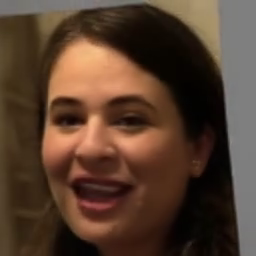}
      \includegraphics[width=0.105\linewidth]{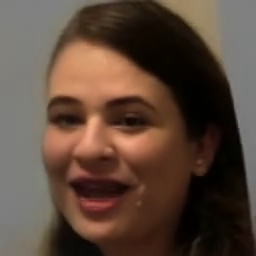}
          \includegraphics[width=0.105\linewidth]{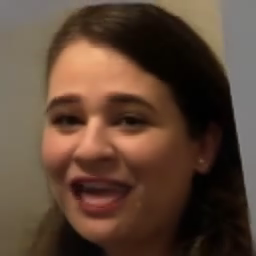}
       \includegraphics[width=0.105\linewidth]{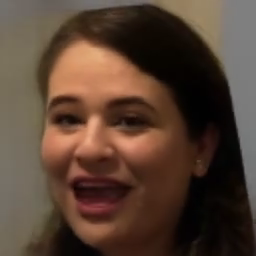}
         \includegraphics[width=0.105\linewidth]{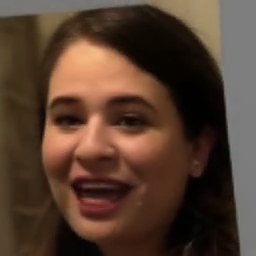}
     \includegraphics[width=0.105\linewidth]{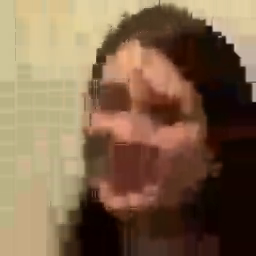}\\
      \includegraphics[width=0.105\linewidth]{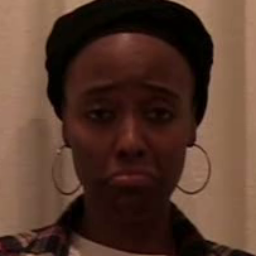}
    \includegraphics[width=0.105\linewidth]{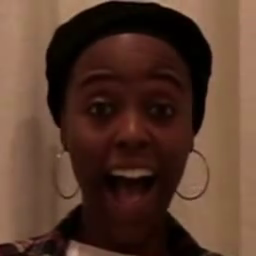}
    \includegraphics[width=0.105\linewidth]{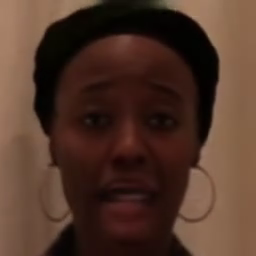}
     \includegraphics[width=0.105\linewidth]{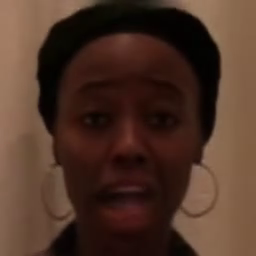}
      \includegraphics[width=0.105\linewidth]{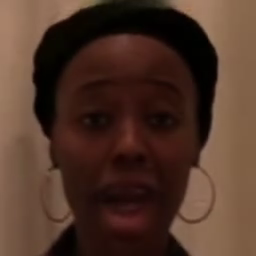}
          \includegraphics[width=0.105\linewidth]{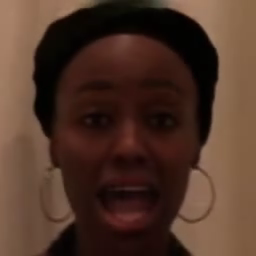}
       \includegraphics[width=0.105\linewidth]{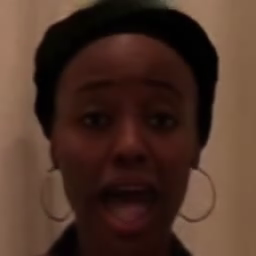}
         \includegraphics[width=0.105\linewidth]{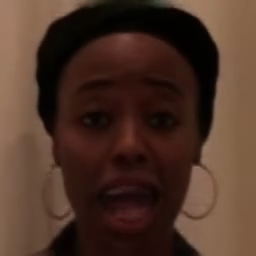}
     \includegraphics[width=0.105\linewidth]{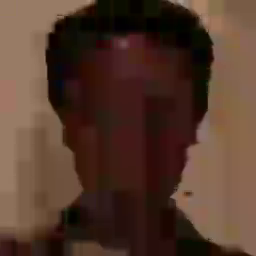}\\
      \includegraphics[width=0.105\linewidth]{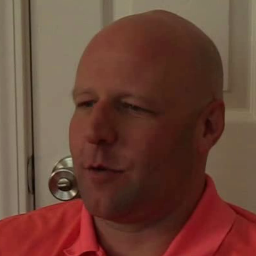}
    \includegraphics[width=0.105\linewidth]{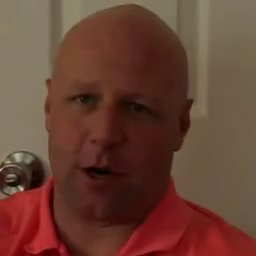}
    \includegraphics[width=0.105\linewidth]{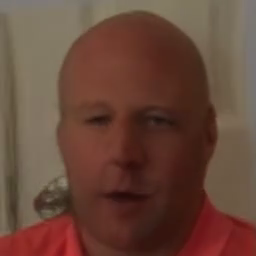}
     \includegraphics[width=0.105\linewidth]{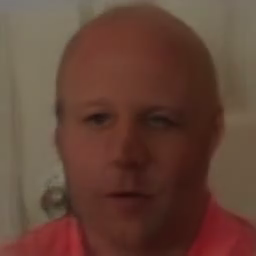}
      \includegraphics[width=0.105\linewidth]{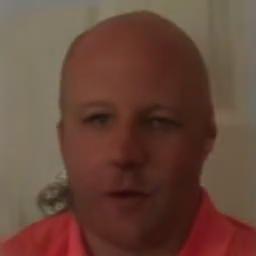}
          \includegraphics[width=0.105\linewidth]{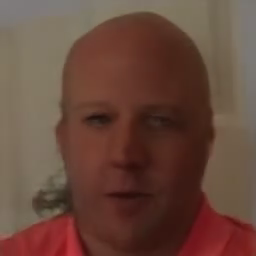}
       \includegraphics[width=0.105\linewidth]{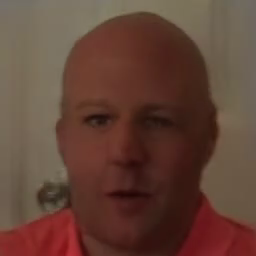}
         \includegraphics[width=0.105\linewidth]{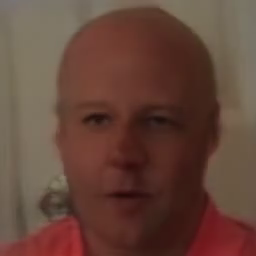}
     \includegraphics[width=0.105\linewidth]{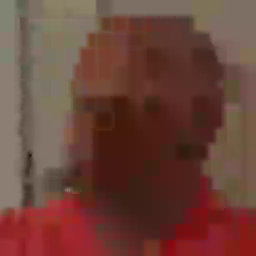}\\
      \includegraphics[width=0.105\linewidth]{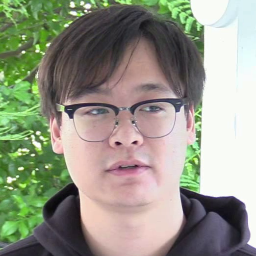}
    \includegraphics[width=0.105\linewidth]{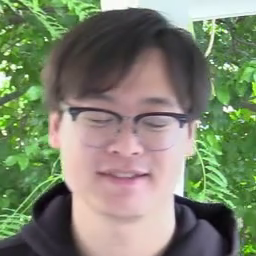}
    \includegraphics[width=0.105\linewidth]{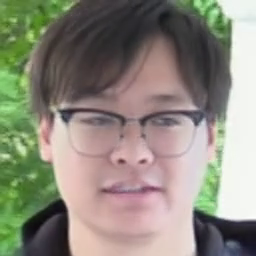}
     \includegraphics[width=0.105\linewidth]{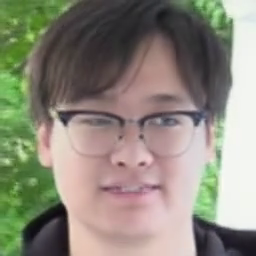}
      \includegraphics[width=0.105\linewidth]{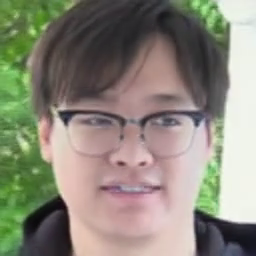}
          \includegraphics[width=0.105\linewidth]{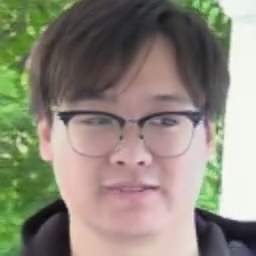}
       \includegraphics[width=0.105\linewidth]{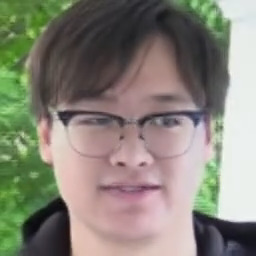}
         \includegraphics[width=0.105\linewidth]{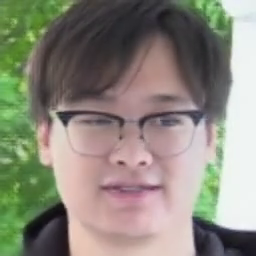}
     \includegraphics[width=0.105\linewidth]{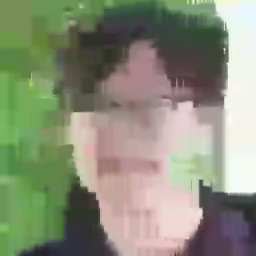}\\
      \includegraphics[width=0.105\linewidth]{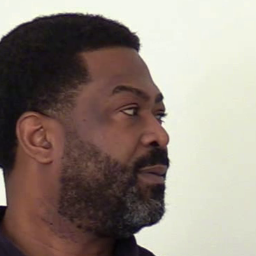}
    \includegraphics[width=0.105\linewidth]{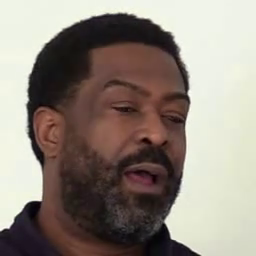}
    \includegraphics[width=0.105\linewidth]{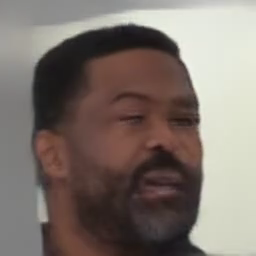}
     \includegraphics[width=0.105\linewidth]{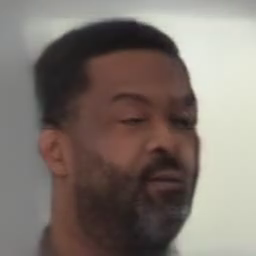}
      \includegraphics[width=0.105\linewidth]{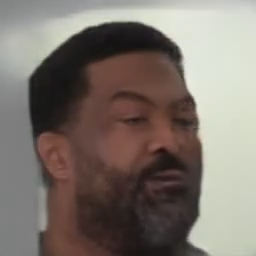}
          \includegraphics[width=0.105\linewidth]{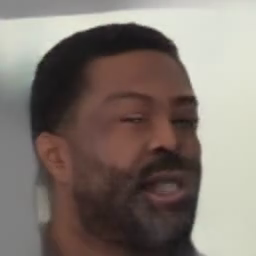}
       \includegraphics[width=0.105\linewidth]{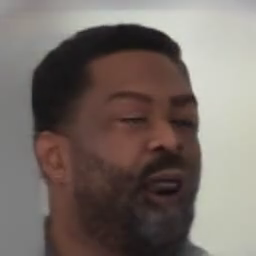}
         \includegraphics[width=0.105\linewidth]{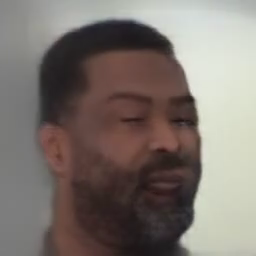}
     \includegraphics[width=0.105\linewidth]{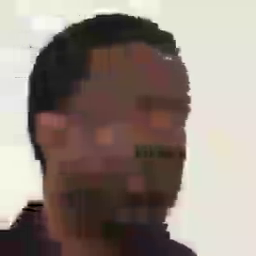}\\
      \includegraphics[width=0.105\linewidth]{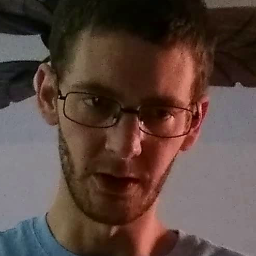}
    \includegraphics[width=0.105\linewidth]{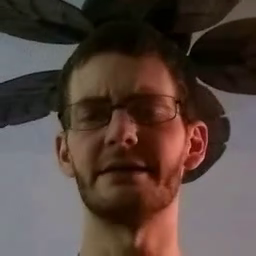}
    \includegraphics[width=0.105\linewidth]{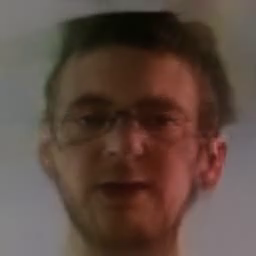}
     \includegraphics[width=0.105\linewidth]{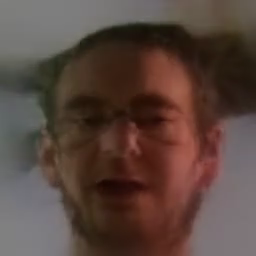}
      \includegraphics[width=0.105\linewidth]{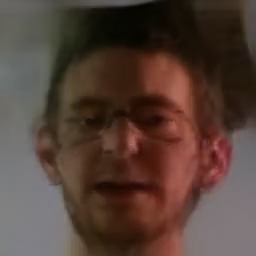}
          \includegraphics[width=0.105\linewidth]{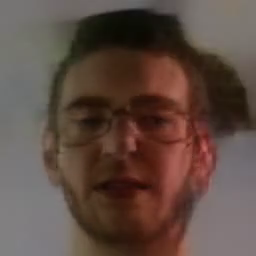}
       \includegraphics[width=0.105\linewidth]{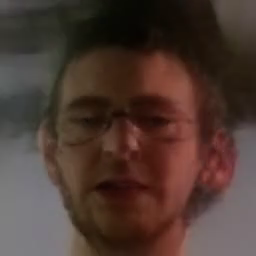}
         \includegraphics[width=0.105\linewidth]{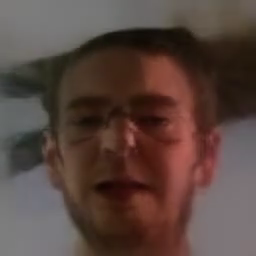}
     \includegraphics[width=0.105\linewidth]{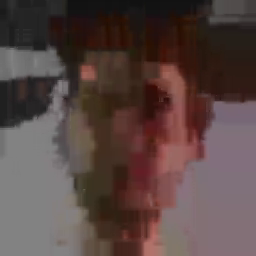}\\
     \begin{tabular}{ccccccccc}
    ~~Sources~~ & ~~~Targets~~~ & ~~MN-10~~ & ~~MN-20~~ & ~MN-10+20 & ~M-SP-10 & M-SP-10+20 & ~M-SP-20~ & H264 9kb/s \\
     \end{tabular}
    \caption{Qualitative results using Motion based variants on mobile architectures, using a $32\times32\times256$ latent space. Each model generates the face given the fixed source frame and the landmarks of the target frame. All models run in real-time on an iPhone 8.}
    \label{fig:qualitativePerfMobile}
\end{figure*}

\subsection{Qualitative evaluation and human study}

Figure \ref{fig:comp} compares results obtained using SegFace, Bilayer, NTH and FOM with adversarial finetuning. We observe skin tones/lightning differences between targets and SegFace results and distortions of personal traits. The Bilayer, NTH and FOM models are qualitatively better. In the last column, we observe a side by side comparison with a Mobile Motion-SPADE model.

Table~\ref{tab:humanStudy} provides a quality assessment of different models by human raters.
Participants are asked to rate images produced by the different models by comparing them in terms of identity and expression preservation, on a scale from 1 to 5. In a first round of evaluations, we display side by side the four main dense models results, and in the second round, Motion Net and Motion-SPADE results using six different mobile architectures. We collect in each case 500 pairwise evaluations, each from five different participants. For dense models results, human scores seem to agree with metrics, ranking FOM first and NTH second. Mobile models results differences are more subtle, but the Hybrid Motion-SPADE using 10 landmarks model seems preferred. The addition of SPADE blocks brings significant improvement in most cases.

\begin{table}[]
    \centering
    \begin{tabular}{ccc}

     \toprule
     \multicolumn{3}{c}{Dense models} \\
     \toprule
               & human identity  & human expression \\
         model &  score & score \\
         \midrule
            NTH & 3.71{\small$\pm$0.041} & 3.77{\small$\pm$0.040} \\
            Bilayer  & 3.70{\small$\pm$0.046} & 3.62{\small$\pm$0.046} \\
            SegFace & 3.11{\small$\pm$0.047} & 3.00{\small$\pm$0.051} \\
            FOM adv & {\bf 3.99}{\small$\pm$0.042} & {\bf 4.00}{\small$\pm$0.041} \\
            \end{tabular}\\
            \begin{tabular}{cccc}
            \toprule
             \multicolumn{4}{c}{Overall human ratings of Mobile models} \\
            \toprule
           & MN-10&    MN-20& MN-10+20 \\
            \midrule
              no SPADE & 3.44{\tiny$\pm0.034$} & 3.40{\tiny$\pm0.034$} & 3.46{\tiny$\pm0.034$}\\
         with SPADE &  {\bf 3.50{\tiny$\pm0.034$}} &  3.46{\tiny$\pm0.035$} & 3.45{\tiny$\pm0.034$} \\
               \bottomrule
    \end{tabular}
    \vspace{2ex}
    \caption{Quality assessment of different dense models on DFDC-50 - Human study. Average scores (Higher is better) with confidence intervals.}
    \label{tab:humanStudy}
    \vspace{-2ex}
\end{table}

Figure~\ref{fig:qualitativePerfMobile} illustrates the quality performance reached on Mobile. The two last lines display challenging cases for the motion based approach. We note that the quality of results degrades in presence of large head rotation. In the last line, a fan on top of the head combined with a bad lightning causes errors in hair reconstruction. Still, the Motion-SPADE results are visually close to the targets, particularly it renders lips and teeth better.
The H264 compression results are displayed given a bandwidth of 9 kbit/s, to be compared to the ones of the Motion-SPADE 20 model that runs the fastest on mobile. This illustrates that at this bandwidth, video transmission is hardly possible using standard codecs, whereas our mobile approach would make the video call possible.

\section{Conclusions}

Our exploration of state-of-the art animation models led us to the following observations:
The Neural Talking head results are qualitatively satisfactory, but the fine-tuning step requirement makes the approach complex to implement in practice.
Using a full face segmentation approach seems unfit to a low bandwidth application.
The Bilayer approach and the FOM methods perform best towards reaching a correct low-bandwidth/quality trade-off on mobile.
Our human study shows that FOM results are preferred in terms of identity and expression preservation.
Focusing on this best candidate approach, we design a novel
 hybrid architecture taking advantage of the high fidelity to the target thanks to the warping principle, and enhancing the quality of important attributes with SPADE blocks.
 Only exploiting polygons induced segments allows this approach to improve quality without high transmission cost. The obtained image quality is close to the one reached by dense models while running in real-time on Mobile CPU. The bandwidth required to send a video is lower than the one required for sending audio.
There are a number of interesting challenges to tackle next to improve quality of the generations, e.g. generating large head rotation movements, hands, or using pupils tracking.

\section*{Acknowledgements}

We would like to thank Elif Albuz, Bryan Anenberg, Stephanie Gu\'aman, Herv\'e Jegou, Stéphane Lathuili\`ere, Aliaksandr Siarohin, Jakob Verbeek and Qiang Zhang for their precious help.

{\small
\bibliographystyle{ieee_fullname}
\bibliography{Arxiv2021}

\begin{thebibliography}{10}\itemsep=-1pt

\bibitem{agustsson2019generative}
Eirikur Agustsson, Michael Tschannen, Fabian Mentzer, Radu Timofte, and Luc~Van
  Gool.
\newblock Generative adversarial networks for extreme learned image
  compression.
\newblock In {\em ICCV}, 2019.

\bibitem{averbuch2017bringing}
Hadar Averbuch-Elor, Daniel Cohen-Or, Johannes Kopf, and Michael~F Cohen.
\newblock Bringing portraits to life.
\newblock {\em Transactions on Graphics}, 36(6):1--13, 2017.

\bibitem{bulat2017far}
Adrian Bulat and Georgios Tzimiropoulos.
\newblock How far are we from solving the 2d \& 3d face alignment problem?(and
  a dataset of 230,000 3d facial landmarks).
\newblock In {\em ICCV}, 2017.

\bibitem{bulat2018super}
Adrian Bulat and Georgios Tzimiropoulos.
\newblock Super-fan: Integrated facial landmark localization and
  super-resolution of real-world low resolution faces in arbitrary poses with
  gans.
\newblock In {\em CVPR}, 2018.

\bibitem{Cao2016RealtimeFacial}
Chen Cao, Hongzhi Wu, Yanlin Weng, Tianjia Shao, and Kun Zhou.
\newblock Real-time facial animation with image-based dynamic avatars.
\newblock {\em Transactions on Graphics}, 35(4), 2016.

\bibitem{cao2018vggface2}
Qiong Cao, Li Shen, Weidi Xie, Omkar~M Parkhi, and Andrew Zisserman.
\newblock Vggface2: A dataset for recognising faces across pose and age.
\newblock In {\em International Conference on Automatic Face \& Gesture
  Recognition}, 2018.

\bibitem{chen2020comprises}
Lele Chen, Guofeng Cui, Ziyi Kou, Haitian Zheng, and Chenliang Xu.
\newblock What comprises a good talking-head video generation?: A survey and
  benchmark.
\newblock {\em arXiv 2005.03201}, 2020.

\bibitem{chen2018fsrnet}
Yu Chen, Ying Tai, Xiaoming Liu, Chunhua Shen, and Jian Yang.
\newblock Fsrnet: End-to-end learning face super-resolution with facial priors.
\newblock In {\em CVPR}, 2018.

\bibitem{Chung18bvox2}
Joon~Son Chung, Arsha Nagrani, and Andrew Zisserman.
\newblock Voxceleb2: Deep speaker recognition.
\newblock In {\em INTERSPEECH}, 2018.

\bibitem{bloggoogle}
Dave Citron.
\newblock ``four new google duo features to help you stay connected", April
  2020.

\bibitem{dai2020fbnetv3}
Xiaoliang Dai, Alvin Wan, Peizhao Zhang, Bichen Wu, Zijian He, Zhen Wei, Kan
  Chen, Yuandong Tian, Matthew Yu, Peter Vajda, et~al.
\newblock Fbnetv3: Joint architecture-recipe search using neural acquisition
  function.
\newblock {\em arXiv 2006.02049}, 2020.

\bibitem{deng2019arcface}
Jiankang Deng, Jia Guo, Niannan Xue, and Stefanos Zafeiriou.
\newblock Arcface: Additive angular margin loss for deep face recognition.
\newblock In {\em CVPR}, 2019.

\bibitem{dolhansky2019deepfake}
Brian Dolhansky, Joanna Bitton, Ben Pflaum, Jikuo Lu, Russ Howes, Menglin Wang,
  and Cristian~Canton Ferrer.
\newblock The deepfake detection challenge (dfdc) dataset.
\newblock {\em arXiv 2006.07397}, 2019.

\bibitem{gafni2019live}
Oran Gafni, Lior Wolf, and Yaniv Taigman.
\newblock Live face de-identification in video.
\newblock In {\em CVPR}, 2019.

\bibitem{Huang_2017_ICCV}
Xun Huang and Serge Belongie.
\newblock Arbitrary style transfer in real-time with adaptive instance
  normalization.
\newblock In {\em ICCV}, 2017.

\bibitem{huff}
David~A. Huffman.
\newblock A method for the construction of minimum-redundancy codes.
\newblock {\em Proceedings of the Institute of Radio Engineers},
  40(9):1098--1101, 1952.

\bibitem{jaderberg2015spatial}
Max Jaderberg, Karen Simonyan, Andrew Zisserman, et~al.
\newblock Spatial transformer networks.
\newblock In {\em NeurIPS}, pages 2017--2025, 2015.

\bibitem{Hyeongwoo2018deepvideo}
Hyeongwoo Kim, Pablo Garrido, Ayush Tewari, Weipeng Xu, Justus Thies, Matthias
  Niessner, Patrick P\'{e}rez, Christian Richardt, Michael Zollh\"{o}fer, and
  Christian Theobalt.
\newblock Deep video portraits.
\newblock {\em Transactions on Graphics}, 37(4), 2018.

\bibitem{korshunova2017fast}
Iryna Korshunova, Wenzhe Shi, Joni Dambre, and Lucas Theis.
\newblock Fast face-swap using convolutional neural networks.
\newblock In {\em ICCV}, 2017.

\bibitem{koufakis1999very}
Ioannis Koufakis and Bernard~F Buxton.
\newblock Very low bit rate face video compression using linear combination of
  {2D} face views and principal components analysis.
\newblock {\em Image and Vision computing}, 17(14):1031--1051, 1999.

\bibitem{lee2020maskgan}
Cheng-Han Lee, Ziwei Liu, Lingyun Wu, and Ping Luo.
\newblock Maskgan: Towards diverse and interactive facial image manipulation.
\newblock In {\em CVPR}, 2020.

\bibitem{liu2020deep}
Dong Liu, Yue Li, Jianping Lin, Houqiang Li, and Feng Wu.
\newblock Deep learning-based video coding: A review and a case study.
\newblock {\em Computing Surveys}, 53(1):1--35, 2020.

\bibitem{liu2015faceattributes}
Ziwei Liu, Ping Luo, Xiaogang Wang, and Xiaoou Tang.
\newblock Deep learning face attributes in the wild.
\newblock In {\em ICCV}, 2015.

\bibitem{lopez1995head}
Ricardo Lopez and Thomas~S Huang.
\newblock Head pose computation for very low bit-rate video coding.
\newblock In {\em International Conference on Computer Analysis of Images and
  Patterns}, 1995.

\bibitem{nagano2018pagan}
Koki Nagano, Jaewoo Seo, Jun Xing, Lingyu Wei, Zimo Li, Shunsuke Saito, Aviral
  Agarwal, Jens Fursund, and Hao Li.
\newblock pagan: real-time avatars using dynamic textures.
\newblock In {\em SIGGRAPH Asia}, page 258, 2018.

\bibitem{Nagrani17vox}
Arsha Nagrani, Joon~Son. Chung, and Andrew Zisserman.
\newblock Voxceleb: a large-scale speaker identification dataset.
\newblock In {\em INTERSPEECH}, 2017.

\bibitem{nirkin2019fsgan}
Yuval Nirkin, Yosi Keller, and Tal Hassner.
\newblock {FSGAN}: Subject agnostic face swapping and reenactment.
\newblock In {\em CVPR}, 2019.

\bibitem{park2019SPADE}
Taesung Park, Ming-Yu Liu, Ting-Chun Wang, and Jun-Yan Zhu.
\newblock Semantic image synthesis with spatially-adaptive normalization.
\newblock In {\em CVPR}, 2019.

\bibitem{waveonevideo2018}
Oren Rippel, Sanjay Nair, Carissa Lew, Steve Branson, Alexander Anderson, and
  Lubomir Bourdev.
\newblock Learned video compression.
\newblock In {\em ICCV}, 2019.

\bibitem{arith}
Jorma {Rissanen} and Glen~G. {Langdon}.
\newblock Arithmetic coding.
\newblock {\em IBM Journal of Research and Development}, 23(2):149--162, 1979.

\bibitem{sandler2018mobilenetv2}
Mark Sandler, Andrew Howard, Menglong Zhu, Andrey Zhmoginov, and Liang-Chieh
  Chen.
\newblock Mobilenetv2: Inverted residuals and linear bottlenecks.
\newblock In {\em CVPR}, 2018.

\bibitem{Santurkar2017generative}
Shibani Santurkar, David Budden, and Nir Shavit.
\newblock Generative compression.
\newblock In {\em Picture Coding Symposium}, 2018.

\bibitem{Siarohin_2019_NeurIPS}
Aliaksandr Siarohin, Stéphane Lathuilière, Sergey Tulyakov, Elisa Ricci, and
  Nicu Sebe.
\newblock First order motion model for image animation.
\newblock In {\em NeurIPS}, 2019.

\bibitem{soderstrom2006very}
Ulrik S{\"o}derstr{\"o}m.
\newblock {\em Very low bitrate facial video coding: based on principal
  component analysis}.
\newblock PhD thesis, Till{\"a}mpad fysik och elektronik, 2006.

\bibitem{son2006ultra}
Le-Hung Son, Ulrik S{\"o}derstr{\"o}m, and Haibo Li.
\newblock Ultra low bit-rate video communication: video coding= pattern
  recognition, 2006.

\bibitem{torres2002proposal}
Luis Torres and Daniel Prado.
\newblock A proposal for high compression of faces in video sequences using
  adaptive eigenspaces.
\newblock In {\em ICIP}, 2002.

\bibitem{tuceryan2000model}
Mihran Tuceryan and Bruce~E Flinchbaugh.
\newblock Model based faced coding and decoding using feature detection and
  eigenface coding, Mar.~28 2000.
\newblock US Patent 6,044,168.

\bibitem{Ustinova2017facehallucination}
Evgeniya Ustinova and Victor~S. Lempitsky.
\newblock Deep multi-frame face hallucination for face identification.
\newblock {\em arXiv}, 1709.03196, 2017.

\bibitem{wan2020fbnetv2}
Alvin Wan, Xiaoliang Dai, Peizhao Zhang, Zijian He, Yuandong Tian, Saining Xie,
  Bichen Wu, Matthew Yu, Tao Xu, Kan Chen, Peter Vajda, and Joseph~E. Gonzalez.
\newblock Fbnetv2: Differentiable neural architecture search for spatial and
  channel dimensions, 2020.

\bibitem{wang2019fewshotvid2vid}
Ting-Chun Wang, Ming-Yu Liu, Andrew Tao, Guilin Liu, Jan Kautz, and Bryan
  Catanzaro.
\newblock Few-shot video-to-video synthesis.
\newblock In {\em NeurIPS}, 2019.

\bibitem{wang2018vid2vid}
Ting-Chun Wang, Ming-Yu Liu, Jun-Yan Zhu, Guilin Liu, Andrew Tao, Jan Kautz,
  and Bryan Catanzaro.
\newblock Video-to-video synthesis.
\newblock In {\em NeurIPS}, 2018.

\bibitem{Wiles_2018_ECCV}
Olivia Wiles, A.~Sophia Koepke, and Andrew Zisserman.
\newblock X2face: A network for controlling face generation using images,
  audio, and pose codes.
\newblock In {\em ECCV}, 2018.

\bibitem{wu2019fbnet}
Bichen Wu, Xiaoliang Dai, Peizhao Zhang, Yanghan Wang, Fei Sun, Yiming Wu,
  Yuandong Tian, Peter Vajda, Yangqing Jia, and Kurt Keutzer.
\newblock Fbnet: Hardware-aware efficient convnet design via differentiable
  neural architecture search, 2019.

\bibitem{yu2018bisenet}
Changqian Yu, Jingbo Wang, Chao Peng, Changxin Gao, Gang Yu, and Nong Sang.
\newblock Bisenet: Bilateral segmentation network for real-time semantic
  segmentation.
\newblock In {\em ECCV}, 2018.

\bibitem{zakharov2020fast}
Egor Zakharov, Aleksei Ivakhnenko, Aliaksandra Shysheya, and Victor Lempitsky.
\newblock Fast bi-layer neural synthesis of one-shot realistic head avatars.
\newblock {\em ECCV}, 2020.

\bibitem{zakharov2019few}
Egor Zakharov, Aliaksandra Shysheya, Egor Burkov, and Victor Lempitsky.
\newblock Few-shot adversarial learning of realistic neural talking head
  models.
\newblock {\em ICCV}, 2019.

\bibitem{zhang2018unreasonable}
Richard Zhang, Phillip Isola, Alexei~A Efros, Eli Shechtman, and Oliver Wang.
\newblock The unreasonable effectiveness of deep features as a perceptual
  metric.
\newblock In {\em CVPR}, 2018.

\end{thebibliography}
}

\section{Appendix}

\begin{figure*}[htb]
     \centering
     \begin{tabular}{c|c}
     \multicolumn{2}{c}{\hspace{-3ex}\includegraphics[width=0.8\linewidth, page=3, clip=True, trim=0.08cm 9cm 2.7cm 0.08cm ]{FigArxiv/all_figs_architecture.pdf}} \\
     \multicolumn{2}{c}{(a) Mobile architectures {\small(dense motion net and landmark extractor)} \vspace{2ex}} \\
     \hline
     \multicolumn{2}{c}{\hspace{-3ex}\includegraphics[width=0.8\linewidth, page=4, clip=True, trim=0.08cm 6.7cm 2.3cm 0.08cm ]{FigArxiv/all_figs_architecture.pdf}} \\
     \multicolumn{2}{c}{ (b) Generator architecture (Input: driving frame, output: generation) \vspace{2ex}} \\
     \hline
    \hspace{-3ex}\includegraphics[width=0.55\linewidth, page=1, clip=True, trim=0.08cm 7.8cm 9.5cm 0.08cm ]{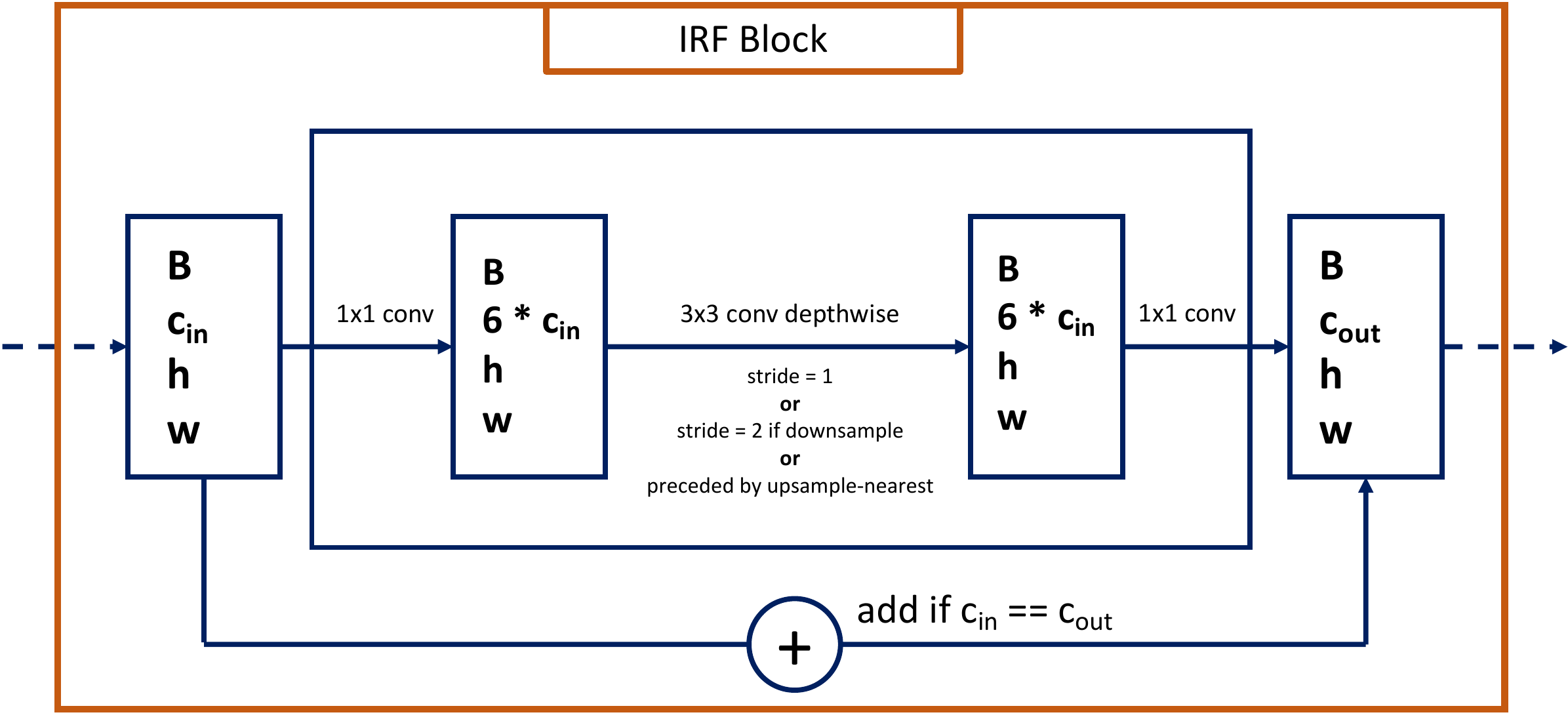} & \includegraphics[width=0.47\linewidth, page=2, clip=True, trim=0.08cm 4.1cm 7.8cm 0.08cm ]{FigArxiv/all_figs_architecture.pdf} \\
    (c) Inverted residual (IRF) blocks & (d) SPADE blocks used in Motion-SPADE
    \end{tabular}
    \vspace{2ex}
    \caption{Schemes of mobiles architectures for the Motion Net ((a), (c)) and Motion-SPADE approaches ((a), (b), (c), (d)).}
     \label{fig:FOM_scheme2}
 \end{figure*}

The details of our mobile architectures are provided in Figure \ref{fig:FOM_scheme2}.

\end{document}